# A Dynamic Data Driven Approach for Explainable Scene Understanding

Zachary A. Daniels[1], Dimitris N. Metaxas[1]

[1]Dept. of Computer Science,
Rutgers, the State University of New Jersey, Piscataway, NJ, USA
{zad7, dnm}@cs.rutgers.edu

**Abstract.** Scene-understanding is an important topic in the area of Computer Vision, and illustrates computational challenges with applications to a wide range of domains including remote sensing, surveillance, smart agriculture, robotics, autonomous driving, and smart cities. In this chapter, we consider the active explanation-driven understanding and classification of scenes. Suppose that an agent utilizing one or more sensors is placed in an unknown environment, and based on its sensory input, the agent needs to assign some label to the perceived scene. The agent can adjust its sensor(s) to capture additional details about the scene, but there is a cost associated with sensor manipulation, and as such, it is important for the agent to understand the scene in a fast and efficient manner. It is also important that the agent understand not only the global state of a scene (*e.g.*, the category of the scene or the major events taking place in the scene) but also the characteristics/properties of the scene that support decisions and predictions made about the global state of the scene. Finally, when the agent encounters an unknown scene category, it must be capable of refusing to assign a label to the scene, requesting aid from a human, and updating its underlying knowledge base and machine learning models based on feedback provided by the human. We introduce a dynamic data driven framework for the active explanation-driven classification of scenes. Our framework is entitled ACUMEN: Active Classification and Understanding Method by Explanation-driven Networks. To demonstrate the utility of the proposed ACUMEN approach and show how it can be adapted to a domain-specific application, we focus on an example case study involving the classification of indoor scenes using an active robotic agent with vision-based sensors, *i.e.,* an electro-optical camera.

## 1 Introduction

Scenes consist of views of real-world environments where surfaces and objects are organized in some meaningful way [1]. Scene understanding is the computational process of perceiving and analyzing scenes based on the spatial and geometric layout of the entire scene and the spatial, functional, and semantic relationships (*i.e.,* the context) that exist between objects within the scene [2]. Another related research topic is image understanding which interprets features, objects, and shapes within an image. Researchers have considered several key tasks in visual scene understanding including segmenting foreground objects from the background (*e.g.,* [3]), identifying the objects present in a scene (*e.g.,* [4]), modeling the relationships between objects in a scene (*e.g.,* [5]), determining the events taking place in a scene (*e.g.,* [6-8]), understanding the layout and structure of the scene (*e.g.,* [9]), classifying the type of scene (*e.g.,* [10]), and tracking objects/targets in a scene (*e.g.,* [11]). Existing approaches for scene understanding are far from perfect. They often make simplifying assumptions about the data, and many of the algorithms are not designed for real-world use cases.

In particular, three problems plague the majority of existing approaches for scene understanding. First, these approaches assume single views that are informative and non-adversarial (*i.e.,* views that are typical of the scene category that do not mislead an agent into deriving mistaken beliefs about the scene) whereas agents placed in real-world environments will often need to consider multiple views, some of which will be uninformative and misleading. Second, these approaches rely on closed set assumptions whereas agents operating in the real-world will frequently encounter new types of scenes consisting of new types of objects arranged in new ways. Thus, real-world models should instead operate under an open set assumption. Third, existing methods for scene understanding utilize black-box models whereas agents in real-world tasks often require transparent models that are capable of explaining their decision-making process.

To illustrate these issues associated with agents that can't handle multiview sensing, open set situations, and scene explanation, consider the most popular datasets for scene classification (*e.g.,* [10, 12-15]). These datasets only consider single views, and the majority of these views are extremely representative of a limited set of known scene categories. Agents operating in real world scenes rarely encounter such pristine data. For example, suppose a robotic agent must explore and analyze indoor scenes. If the robot is randomly placed within the environment, it might encounter sub-optimal views that are either 1) uninformative (*e.g.,* the robot is facing a blank wall) or 2) adversarial (*e.g.,* the robot is looking through an opening into an adjacent room belonging to a different class of scenes). In order to make better-informed decisions, it is important to imbue agents with the ability to intelligently explore and reason about complex environments.

Another problem with existing approaches for scene understanding is that the agent/model is assumed to be fully equipped with all relevant knowledge about the types of scenes it might encounter; *i.e.,* existing algorithms for scene understanding typically make closed set assumptions (*i.e.,* the agent assumes that when it is deployed, it will only encounter situations similar to those it encountered during its "training" period when it learned/constructed its internal models). In real-world applications, closed set assumptions are rarely satisfied. On occasion, agents will encounter novel scene categories, events, and arrangements of objects. When met with a new scene, instead of operating on false premises and making misleading and incorrect decisions, an agent should have the capacity to reject making a decision and/or request feedback from a human partner. The human can then validate that the encountered scene is atypical with respect to the agent's knowledge base and subsequently collect additional training data specifically about the new scene category. The new data would then be fed into the agent which would use the data to update its internal knowledge and machine learning (ML) models, enabling the agent to operate under open set assumptions [16, 17] and perform continuous learning.

Existing approaches for scene understanding rely on complex, black-box models whereas many safety-critical applications require models that can easily be interpreted and debugged. As an example in the computer vision domain, these models are typically convolutional neural networks (CNNs) [18], which have such a large number of parameters that humans cannot easily track and understand the CNN decision-making process [19, 20]. There are many applications of scene understanding where it is extremely important that the models can 1) generate explanations to help support decisions that must ultimately be made by humans (*e.g.,* military and healthcare applications where the wrong decision can, for example, result in unnecessary death) and 2) be debugged (*e.g.,* autonomous cars where the limitations of the model must be understood extremely well or else passenger and pedestrian safety is at risk).

In this chapter, we introduce the problem of active explanation-driven scene understanding based on sensor data, and thus model more realistically real-world applications of situational awareness. Suppose there is an agent with one or more sensors attached. The agent is placed in some unknown environment, and based on its sensory input, the agent must assign some label to the perceived scene. The agent can adjust its sensor(s) to capture more details about the scene, but there is some cost associated with sensor manipulation, and as such, it is important for the agent to understand the scene in a fast and efficient manner. It is also important that the agent understands not only the global state of a scene (*e.g.,* the category of the scene or the major events taking place in the scene) but also the local state characteristics/properties of the scene that support decisions and predictions made about the global state of the scene. In essence, the agent should not just map input sensor data directly to some prediction but instead be capable of generating explanations for why certain predictions and decisions were made and/or why a certain action was taken. Finally, when the agent encounters an unknown scene category, it must be capable of 1) refusing to assign a label to the scene, and instead, request aid from a human, and 2) updating its underlying knowledge base and ML models based on feedback provided by the human.

In the following research, we define the problem of active explanation-driven scene understanding, and we present a set of methods and the ensuing framework for solving this problem, based on the Dynamic Data Driven Applications Systems (DDDAS) paradigm [21] wherein there is a feedback control loop between sensor manipulation and (dynamic) data driven modeling. We call our framework ACUMEN: Active Classification and Understanding Method by Explanation-driven Networks. In our ACUMEN framework, there is view-level, scene-level, and human-level knowledge processed as:

1. The agent extracts a set of atomic, human understandable properties (*i.e.,* view-level knowledge) about a scene based on one or more sensors.
2. The view-level knowledge is fused with any existing scene-level knowledge discovered from prior exploration.
3. Based on the scene-level knowledge, the agent determines if some label that describes the global state of the scene (*e.g.,* the class of the scene or the major event taking place in the scene) can be assigned to the scene with high confidence. If it can, the agent reports its prediction(s), requests verification from a human partner, and if the decision is approved by the human partner, ends exploration.
4. Otherwise, the agent must decide whether the scene is atypical (*i.e.,* out-of-distribution) or if its current knowledge of the global state of the scene (*i.e.,* the scene-level knowledge) is insufficient.
5. If the scene-level knowledge is insufficient, the agent determines the next best view based on its existing knowledge, adjusts its sensor(s), and steps 1-4 are repeated.
6. If the agent determines the scene is atypical, it requests feedback from a human. The human verifies the agent's assessment, collects additional data about the new type of scene, and the agent updates its internal knowledge representation and ML models, and ends exploration.

## 2  Example Domains and Applications

Active explanation-driven scene understanding is an important computational problem with applications to a wide range of domains including remote sensing, defense, surveillance, robotics, and autonomous driving. To illustrate the effectiveness of our ACUMEN approach, in this section, we briefly discuss examples of how the scene understanding problem is manifested in some of these domains; specifically, we present the example cases of remote sensing, surveillance, and robotics. For each example application, we discuss:

1. What is the agent?
2. What are the sensors?
3. What is the goal of the agent?
4. What information about the scene is useful for making informed decisions about which actions to take and generating explanations for why specific actions were taken?
5. What actions can be taken by the agent?
6. What are the costs associated with taking these actions?

### 2.1  Remote Sensing

Considering an example problem in remote sensing, suppose the agent is an unmanned aerial vehicle (UAV) with sensors consisting of electro-optical or thermal cameras. The Unmanned Aerial System (UAS) goal is to survey some land mass in order to make some decision about the utility of the land; *e.g.,* finding a plot of land that is suitable for growing a specific type of crop. The agent needs to acquire information about the land mass of interest based on its sensor data. The agricultural information could include properties such as the color and type of soil, the color and amount of vegetation, the land's susceptibility to flooding, the average temperature of the land, and the average of amount of direct sunlight. Based on these properties, the agent can take several actions. It can make a prediction about whether the land is suitable for the given crop and return to base; it can discern that there is something unusual about the land (*e.g.,* maybe the land is anomalous due to a rockslide) and request human intervention and feedback; or it can determine it does not have enough information and adjust its sensors (*e.g.,* focal length) or trajectory (*e.g.,* by manipulating the path of the vehicle). For sensor managment, the cost of making a sensor adjustment correlates to the amount of fuel needed to move the vehicle to a specific location, and the cost of making a prediction is correlated to the cost associated with further testing of the land (*e.g.,* by collecting soil and performing soil tests and ground-level surveying). Ultimately, the agent should provide information useful for determining profits or losses from growing a given crop.

## 2.2 Defense

In homeland security and military applications active explanation-driven scene understanding is a significant challenge with dynamic and complex scenes, *e.g.,* for intelligence, surveillance, and reconnaissance (ISR). The agent might be an aerial vehicle (UAV). Its sensors may include electro-optical cameras, thermal cameras, a radar system, or a laser system. A goal of the agent could be to determine if some region is controlled by enemy forces. The agent needs to extract human-understandable concepts about the region such as the presence of different vehicles (types of, number, size, and density); buildings (occupied, large, construction), the presence of weapon systems, and people (cultural mix, the density of crowds, and presence and number of bystanders). There are several actions the agent can take. It can generate a report of how likely the region is to contain an enemy base and forward this report to humans who will then ultimately make a decision with respect to the next steps to be taken (*e.g.,* military action, additional surveillance, ignoring the region if no threat is deemed present). It is vitally important that the agent can explain its recommendation because the consequences of taking an incorrect action based on erroneous information is a matter of life-and-death. The agent may adjust its sensors by adjusting its path/trajectory. It may determine that there is something atypical about the scene and request feedback from human operators/partners (and in the defense domain, such abnormalities are extremely important). The cost associated with sensor manipulation is related to the cost of the fuel needed to move the vehicle to a specific location, but it also relates to time spent in the air because this correlates to the risk of being identified by enemy troops. The cost associated with reporting results to the relevant human decision-makers relates to the cost of the next actions taken which can be a matter of life-and-death if military action is taken or a threat is incorrectly ignored.

## 2.3 Surveillance Using Sensor Networks

A network of surveillance cameras can determine whether a crime or terrorist attack is taking place. Instead of having to simultaneously analyze video feeds from hundreds of cameras, the system would make fast decisions about whether police intervention is necessary by focusing on a few key cameras. The agent, in this case, is the sensor network. The goal is to highlight the minimal number of video feeds that are most important for determining whether or not a crime is taking place. Each camera should capture human understandable properties of the scene such as crowd density, the movement patterns of people, and the presence of violence or weapons. The agent must be capable of acting by making a prediction about whether a crime is occurring, and subsequently, it must identify the relevant camera feeds, and present these camera feeds to security or police officer(s) while also highlighting the rationale as to why these views are important using the human-understandable properties. The labeled camera feeds would enable the officer(s) to quickly evaluate the situation and decide what further action to take. Alternatively, the agent could see something anomalous in one of its camera feeds and request immediate intervention and feedback from a human companion. Finally, the agent could decide that the current video feed(s) that it is evaluating is insufficient for making a decision about whether or not a crime is occurring, and as a result, it would need to decide which camera feed(s) should be examined next. The cost of making a sensor adjustment (*i.e.,* changing camera feeds) is related to time which correlates with the likelihood that the criminal may escape or additional harm that could be incurred by bystanders. The cost of making a decision relates to the tradeoff between the potential for preventing future harm and incorrectly wasting resources on a false alarm as well as the future costs associated with ignoring a crime that should have been pursued. Thus, the cost is related to safety and security.

### 2.4 Robotics

There are also many applications in robotics that fit into the paradigm of active explanation-driven scene understanding. For example, a robotic agent may be randomly placed in some environment, and the agent must localize itself with respect to some map. The agent can extract human-understandable properties of the scene (*e.g.,* for indoor scenes, such properties might include the presence of specific objects in the scene, the dimensionalities of a room, the colors of the walls, the type of flooring, *etc.*) using camera properties (*e.g.,* RGB, depth, *etc.*). The agent can take several actions. The agent can make a prediction about its location, decide that it is completely lost (*e.g.,* if the map provided to the agent is incorrect or the environment has changed since the creation of the map) and request aid from a human partner, or determine that it doesn't have enough information about the environment, and subsequently adjust its sensors (*e.g.,* by moving its body, sensors, zoom). The cost of making a prediction about its location depends on the specific application the robotic agent is trying to accomplish (*e.g.,* a robot vacuum cleaner vs a search-and-rescue robot). The cost of adjusting its sensors is associated with factors such as energy use and wear-and-tear on the hardware.

The aforementioned applications demonstrate the impact and utility of our DDDAS-based approach for scene understanding. To demonstrate the concept and validate the functionality and utility of our framework, we focus on a simple use and the more constrained problem of "active explanation-driven classification of indoor scenes" (introduced in Section 5).

## 3 Related Work

Before discussing the details of our DDDAS-based methods and the ensuing framework and before discussing how these apply to the case study application of "active explanation-driven classification of indoor scenes", we first outline related areas of research and present our work in relation to some alternate state-of-the-art approaches in related application areas.

### 3.1 Active Vision and Active Learning

The work introduced in this chapter is closely related to the *active vision* paradigm [22-26] where the task of visual perception is treated as a dynamic and purpose-driven process, and the imaging sensors are controlled by an active observer. The work also touches on the subject of *active learning* [27, 28] where a ML algorithm must efficiently and intelligently query a user (or other sources of information) about the labels of unknown data points. The DDDAS paradigm incorporates elements of both active vision and active learning: an agent imbued with DDDAS capabilities can adjust both its sensors and the learning models for improved decision-making, resulting in a feedback loop between sensor manipulation and dynamic data driven modeling.

Our ACUMEN approach shares similarities with work on active learning for scene classification (*e.g.*, [29-32]). These works consider scene classification in an active setting where the models must determine which unlabeled samples are likely to be most informative for training or updating classifiers in order to improve performance on the scene classification task. In contrast, our work also considers scene classification in active settings, but it focuses on 1) actively selecting the most informative views for making a scene classification decision and 2) actively updating the underlying ML models. The work presented in this chapter does not consider how to select the most informative training samples for scene classification. Of the aforementioned works in applying active learning to scene classification, Li and Guo's method [31] is most similar to ours. In Li and Guo's work, object-based features are used for explainable scene classification, and an active learning component is introduced in order to improve the model based on unexpected scenes. However, unlike our approach, Li and Guo's method relies on classic methods for feature extraction and only operates on data consisting of clean, single views (*i.e.,* there is no exploration of the scene from multiple views).

Our work is also similar to the active scene recognition task proposed by Yu *et al.* [33]. In Yu *et al.*'s method, object-based high-level knowledge is used to actively guide the attention of a ML model in scene images and videos for improved scene classification. As with Li and Guo's method method, the approach by Yu *et al.* relies on traditional hand-engineered visual features and assumes clean, single views of images and video. Additionally, the approach by Yu *et al.* does not address the case where unknown scenes are encountered.

Reineking, Schult, and Hois [34] propose another method for actively exploring scene images using object-based features. Their approach utilizes knowledge derived from a domain ontology and a statistical model in order to analyze scenes based on recognized objects. They propose an active vision-based framework whereby different object class detectors are applied to the current scene based on domain knowledge according to the principle of maximum information gain. Evidence from the object detectors is combined in a belief-based framework in order to classifiy a given scene. Once again, Reineking, Schult, and Hois' method predates methods for automatically learning feature representations and only considers single highly-informative views. In addition to the work discussed above, a number of other efforts attempt to merge active learning/vision with scene classification in novel ways (*e.g.,* [35, 36]).

Methods based on active vision have also been used to address other problems related to scene understanding. For example, methods have been proposed for active scene exploration (*e.g.,* [37, 38]), viewpoint selection (*e.g.,* [39-42]), and active object localization and recognition(*e.g.,* [43-47]).

**3.2 High-Level Information Fusion**

Our approach is related to the problem of high-level information fusion (HLIF) (*e.g.,* see [48-51]). Generally, low-level IF includes object detection, while HLIF is focused on situation assessment and user coordination. HLIF involves fusing information captured by multiple sensors based on high-level symbolic, semantic, and syntactic information. Information captured by individual sensors can be noisy and incomplete. The goal of HLIF is to combine information captured from multiple sensors, knowledge bases, and contextual models in order to reduce noise in the information captured by any of the individual sensors while simultaenously extracting richer information that more completely describes some environment. Our approach follows this paradigm by making predictions about the global states of scenes based on symbolic semantic information captured from different views obtained by manipulating a set of sensor(s). Hence, the scene corresponds to the HLIF notion of a situation.

**3.3 Open Set/World Recognition**

In our approach, an agent must be capable of automatically understanding when new, previously unseen situations are encountered. This involves constructing models capable of quantifying uncertainty in their predictions and operating based on an open set assumption. Open set recognition (*e.g.,* [16, 17, 52-57]) is a task where a given classification model can either 1) assign a label from a known closed set of labels, or 2) if confronted with an instance of a new class, reject making a decision about the class assignment and flag the instance as belonging to a new class. Several methods go one step further and are capable of incrementally updating their underlying ML models as new classes are encountered (*e.g.,* [17, 55, 58-64]). Our approach is compatible with most existing algorithms for open set recognition; however, our model utilizes open set recognition in a new way by updating both the underlying ML models and the underlying knowledge representation when new classes are encountered.

### 3.4 Explainable Models for Visual Recognition Tasks

Our approach is concerned with explaining complex ML models. Many methods have been proposed for improving the post-hoc explainability of complex learning-based visual recognition models. In particular, a lot of work has been done on trying to explain the decisions of already trained deep neural networks [65-71]. These approaches generally involve trying to find the input that maximally activates the neurons related to some decision (*e.g.,* using activation maximization [72]), quantifying the interpretability of the internal latent representations of deep neural networks via network dissection [73], and identifying the pixels and edges in an image that are most informative with respect to a model's final decision [74-80]. Many of these methods produce explanations highlighting *which* regions of an image are important for the model's decision without explaining *why* the regions are important. Other methods (*e.g.,* [81-92]), including our approach, explicitly ground the decisions of complex ML models to human-understandable concepts. In contrast to existing approaches, which are designed to operate on single images in static environments, our work focuses on explainable models for tasks related to dynamic scene understanding.

### 3.5 Similar Dynamic Data Driven Applications Systems

The DDDAS paradigm has been previously applied to a wide range of applications involving understanding some type of "scene" using various types of sensors. This section provides a few motivating examples of scene understanding-related applications which utilize DDDAS. In contrast to the approach discussed in this chapter, which proposes a general framework for applying DDDAS to explainable scene understanding, previous works were designed for specific applications.

In the intelligence, surveillance, and reconnaissance (ISR) realm, DDDAS has been applied to problems such as automatic target recognition and tracking (*e.g.,* [11, 93-103]) and situational awareness (*e.g.,* [104-111]).

Similarly, we can consider applications in remote sensing settings. One application in this domain is applying DDDAS to weather forecasting and wildfire detection and prediction (*e.g.,* [112-117]). Another application utilizing DDDAS in a remote sensing setting is remotely assessing water quality (*e.g.,* [118]) and similarly, remotely monitoring and managing oil spills (*e.g.,* [117, 119]).

If we loosely define the concept of a "scene", we can consider many other applications. For example, DDDAS has been applied to managing, simulating, and planning construction operations (*e.g.,* [120-122]), which involve understanding the activities of construction equipment and understanding how different pieces construction equipment interact with one another and their environment. Similarly, DDDAS has been applied to transportation modeling and traffic management (*e.g.,* [123-125]), which involve understanding how different vehicles interact with one another and their environment. Finally, we can consider many other problems involving modeling various environmental systems including meteorological, oceanographical, hydrological, and geographical (see [126] for an overview and review of such applications and the DDDAS methods proposed for addressing them).

Despite each of these applications being very different from one another, they all involve analyzing, understanding, and making predictions and inferences about some type of "scene" (loosely-defined from single situation, place of occurrence, or element in a sequence) based on analyzing measurements output by some type of sensor which can be manipulated.

## 4 A Dynamic Data Driven Framework for Active Explanation-Driven Scene Understanding

In this section, we describe the overall framework using DDDAS for explainable scene understanding. In Fig. 1, we show a flowchart of our approach, where the "representation model" captures information and knowledge about the semantic content of scenes, and the prediction model relates the semantic content to actions and decision-making. The architecture is split into four stages:

**Stage 1 – Sensing:** Data is collected from some set of sensors about a single "view". We define a view as the set of sensor measurements at a single point in time focused on a specific region of interest; *e.g.,* if the sensor is a camera, a view would consist of a single image capturing a specific portion of a scene.

**Stage 2 – Processing:** The raw sensor data are processed into a form that is human-understandable. The processing stage consists of two parts. *First*, it involves using a model to map the view-level sensor measurements to semantic features. For example, this might involve using a convolutional neural network (CNN) to map from an image (represented as a matrix of pixels) to some visual attributes (properties that describe a visual scene; *e.g.,* when analyzing an outdoor scene, visual attributes might include the following: weather::cloudy, foliage::brown, has_lake::true, *etc.*). *Second,* the processing involves fusing information from multiple views in order to extract semantic information at the *scene*-level; *i.e.,* the processing stage requires performing HLIF as discussed in the Related Work section. As views are sequentially encountered, new sensor measurements are obtained, and these measurements are converted to view-level semantic information, and the scene-level semantic information is updated.

**Stage 3 – Decision-Making/Prediction:** At every time step (*i.e.*, whenever encountering a new view), the system agent needs to make some decision about what action should be taken next. These decisions must be made based on the scene-level semantic information that have been collected up to the current time step. One advantageous method uses a model with scene-level semantic information as features to feed a ML model that predicts one of three actions: 1) with high confidence, assign some label to the scene (*e.g.,* the category of the scene or actions taking place in the scene), 2) determine that the scene is atypical given the agent/model's past experiences, or 3) determine that no prediction can be made with high confidence, and as a result, the sensor must be adjusted to collect additional information/evidence for making a decision.

**Stage 4 – Updating:** The agent can update itself in two ways. *First,* if an atypical type of scene is encountered (*e.g.,* if the scene belongs to a category that has never been encountered before), then the model must be updated to account for this new information. In our framework, the agent must update its internal ML models and knowledge representations when it encounters atypical scenes. To do so, the agent queries a human to verify that the scene is truly atypical. If the agent made a mistake and the new scene instance belongs to a known scene class, the new instance can be treated as an additional training instance to update the agent's models and representations when they are next updated. If the new instance belongs to an unknown scene class, the model requests additional data curation from a human (*i.e.,* training instances) about the new scene class, and the agent updates its ML models and knowledge representations. In the *second* way, an agent can update itself is by adjusting/manipulating its sensors when it determines there is not enough information to make a decision to output a prediction about the scene class or reject making any prediction about the scene class when atypical scenes are encountered. To do so, the agent can consider computational models based on reinforcement learning.

Up to this point, we have not specified the technical details of how our DDDAS framework for explainable scene understanding works. This is because the framework is designed to be very flexible with respect to the types of problems it can address. In the next section, we will examine how our framework can be applied to a specific use case: the active, explanation-driven classification of indoor scenes. In particular, we will offer specific technical details about the problem formulation and the ML models and knowledge representation employed by our prototype system for this application.

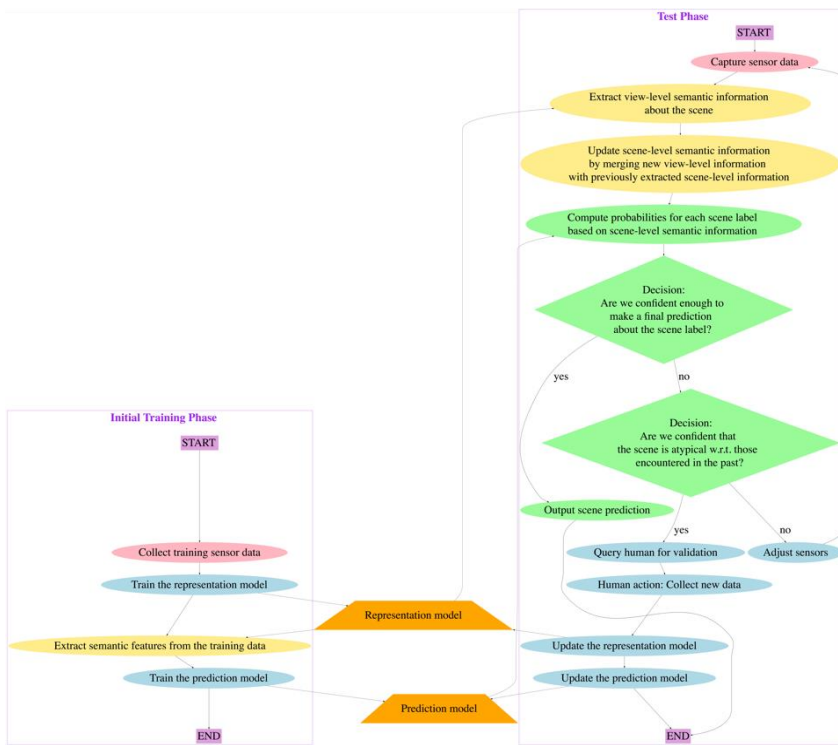

**Fig. 1.** An overview of our DDDAS-based framework for explainable scene understanding.

## 5 Case Study: Active Explanation-Driven Indoor Scene Classification

### 5.1 Problem Overview

Consider the problem of the *active explanation-driven classification of indoor scenes*. It is assumed that an agent is placed in the center of an indoor space and with few sensor adjustments, must assign a category to the scene (*e.g.*, kitchen, dining area, *etc.*). After the agent is situated, it 1) captures an image, 2) based on the captured image, extracts relevant human-understandable semantic information about the scene, and 3) using the semantic information must make a *grounded* decision about which action to take next. The agent can 1) assign a label to the scene from a known set of labels, 2) adjust the orientation of its camera to gather more information about the scene, or 3) determine that the scene is unlike any its seen before and request more information from humans. If the agent adjusts its sensor, it must be capable of fusing existing information with newly obtained information. If the agent identifies a new type of scene, it must be able to 1) update its existing knowledge base, and 2) update its visual recognition models.

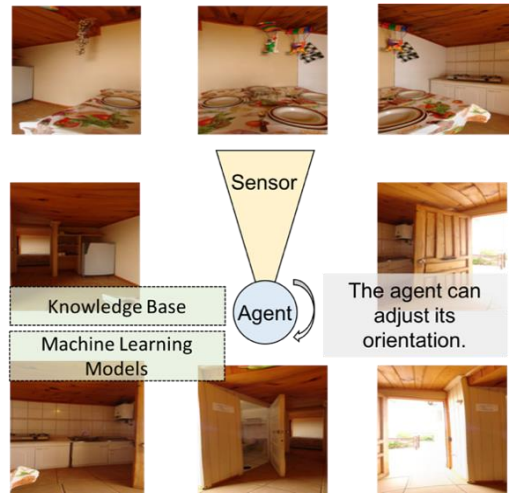

**Fig. 2.** The problem setup for the active explanation-driven classification of indoor scenes

### 5.2 Sensing: Understanding the Input Data

The sensor being used in our experiments consists of a standard electro-optical camera capable of capturing RGB images (see Fig. 2). Specifically, the agent is capable of capturing individual RGB images of single views once per time step (*i.e.*, every time an action is taken). To thoroughly evaluate our method in a controlled manner, we simulate an active agent with a single camera that is capable of pivoting on some central axis where the agent can capture views at 8 positions spaced at even intervals as shown in Fig. 2. To collect the sensor and label data needed to simulate such an agent, we extend the SUN360 dataset [127] of *panoramic* scene images. The SUN360 dataset enables simulating an agent that can manipulate a camera in order to obtain different views. From the dataset, 14 common indoor scene categories in a residential setting were selected: {atrium, bathroom, bedroom, child's room, church, classroom, conference room, dining room, kitchen, living room, office, restaurant, theater, and workshop}. 35 instances for each scene category are annotated. For each instance, we select eight views at evenly spaced intervals. Thus, the dataset consists of a total of 3,920 images. Furthermore, for each image, the presence or absence of 316 unique object classes (*e.g.*, bed, chair, lamp, *etc.*) is recorded. In our experiments, only 201 objects are considered that appear in at least five images. These objects will serve as the foundation for the semantic features needed to generate explanationswhen making predictions and deciding which action to take. In our experiments, for each category of scene, 20 scene instances are used for training (160 images), five scene instances for validation (40 images), and 10 scene instances for testing (80 images), for a total of 280 training scenes (2,240 images), 70 validation scenes (560 images), and 140 test scenes (1,120 images).

**5.3 Processing: From Pixels to Scenarios**

Next, the raw images captured from the camera are processed in order to extract semantically-meaningful representations that can be understood by humans as well as features for the ML models that will ultimately be responsible for deciding which action should be taken next. The processing stage involves three parts: 1) defining the semantically-meaningful human-understandable representation, 2) learning a model to map from the raw sensor data (images) to the aforementioned representation, and 3) fusing the view-level semantic information in order to extract scene-level information.

**5.3.1 The Scenario: A Dynamic Data Driven, Human-Understandable Representation for Scene Understanding**

First, there is a need to specify a human-understandable, semantically grounded representation that is 1) discriminative for scene classification and 2) capable of being accurately extracted from visual data.

One sensible representation is using the presence of objects as features for scene classification. Objects are an intuitive representation for scene understanding and especially for scene classification. In a residential setting, by knowing that there is a refrigerator and oven in a scene, it is easy to hypothesize that the scene is a kitchen, and by knowing there is a table with dinner-plates and utensils (fork, spoon, knife) in a scene, it is easy to hypothesize that the scene is a dining room or a kitchen that includes a dinning area. To validate the hypothesis that objects are a powerful representation for scene classification, we run a simple experiment. We train a simple linear multinomial logistic regression model using the *ground-truth* presence of objects as a binary feature vector using the data discussed in the previous section. As a baseline for comparison, we consider purely visual features which are extracted using a ResNet-18 model [128], a popular CNN architecture that has been shown to perform well on visual recognition tasks, fine-tuned on individual images from our dataset. In our experiments, we examine both single-view performance and all-view performance (where we make decisions based on all eight views for a scene). For the ResNet-18 model, the agent naïvely fuses information from all views by outputting the scene category which has the maximum predicted probability based on the maximum individual view prediction probability. Results are reported in Table 1. Results suggest object presence is a powerful representation for scene classification, significantly outperforming the visual features on both the single-view and multi-view classification tasks in terms of accuracy.

**Table 1.** Object-based representations for scene classification are very discriminative, especially when compared to purely visual features.

| Method | Single-View Accuracy | All-View Accuracy |
| --- | --- | --- |
| Standard ResNet-18 CNN | 0.504 | 0.586 |
| Ground Truth Object Presence + Logistic Regression | 0.654 | 0.793 |

In practice, an agent does not always have access to ground truth object information, and instead the object presence information must be estimated from visual data. Fundamentally, there is a need to understand how well each object can be recognized from data. In order to perform multi-object recognition, we train a ResNet-18 CNN to simultaneously predict the presence of all objects in an image (a multi-label classification problem). For this experiment, the training utilized a weighted binary multi-label cross-entropy as the loss function. By examining the performance of this model, there are some problems when using object presence as features for scene classification. Fig. 3 shows the average precision for the predictions of each object in the dataset. The average precision is a single number that approximates the area under the precision-recall curve. Ideally, the average precision should be close to 1. In contrast, Fig. 3 shows that most objects in the dataset cannot be accurately determined from visual data. This is likely because most objects are very small and have very few training examples, so the CNN cannot learn visual patterns capable of capturing all of the deformations and variations of appearance for the majority of objects, and thus, the model cannot achieve good generalization performance when presented with new data. Furthermore, if the predicted object probabilities are used as features for scene classification, the model can no longer be considered explainable because the features can no longer be trusted. Similarly, in Table 2, we see that scene classification performance diminishes when noisy predicted objects are used as features for scene classification.

**Table 2.** Classification performance diminishes when noisy predicted object presence features are used for scene classification.

| Method | Single-View Accuracy | All-View Accuracy |
| --- | --- | --- |
| Ground Truth Object Presence + Logistic Regression | 0.654 | 0.793 |
| Predicted Object Presence + Logistic Regression | 0.473 | 0.564 |

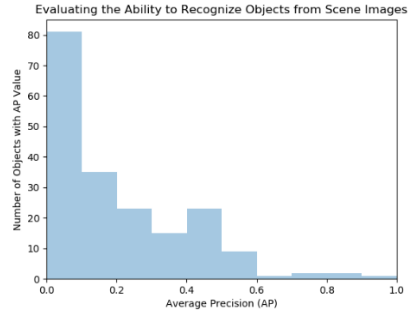

**Fig. 3.** Results show that (for our dataset) the vast majority of objects are unable to be accurately recognized from scene images.

Both of the aforementioned results suggest that using raw object presence features is not ideal for the task of explainable scene classification. Instead, the present work pursues a new interpretable and dynamic data driven representation which we term: the *scenario and a neural network capable of identifying and recognizing scenarios from visual data termed the ScenarioNet* [84]. Scenarios are based on *sets of frequently co-occurring objects* and satisfy a number of key properties:

1. Scenarios are composed of one or more objects.
2. The same object can appear in multiple scenarios, and each scenario should capture the context in which the object appears; *e.g.,* {keyboard, screen, mouse} and {remote control, screen, cable box} both contain screens, but in the first scenario, the screen represents a computer monitor, and in the second scenario, it represents a television screen.
3. Scenes can be decomposed as a union of scenarios; *e.g.,* an instance of a dining room might decompose as: {table, table cloth} ∪ {plate, cup, fork, spoon, knife} ∪ {chandelier, vase, flowers}, and
4. Scenarios are robust to missing objects because scenarios can be present in a scene without all of their constituent objects being present.

Instead of predicting the presence of individual objects in a scene, our model predicts the presence of scenarios instead, which can be recognized with much higher accuracy. This allows the model to avoid recognizing only individual objects and instead recognize the context that exists between objects, a common idea in computer vision (*e.g.,* see [5]), resulting in a model which outputs less fine-grained but more trustworthy explanations.

Scenarios can be identified from data using Pseudo-Boolean Matrix Factorization (PBMF), as in Figure 4. PBMF takes a binary object-scene matrix $A$ and decomposes it into 1) a dictionary matrix $W$ where each basis vector is a scenario and 2) an encoding matrix $H$ that expresses a scene instance as a union of scenarios.

We formulate PBMF as the following:

$$\min_{W,H} P0 + \alpha_1 * P1 + \alpha_2 * P2 + \alpha_3 * P3 + \alpha_4 * P4$$

$$s.t.\ W \in [0,1]^{m \times k}, H \in [0,1]^{k \times n},$$

$$\Omega_{ij} = \max\left(A_{ij} * \left(1 + \log\left(\frac{N_{instances\_all}}{N_{instances\_object}}\right)\right), 0.5\right),$$

$$P0 = ||\Omega \bullet (A - \min(WH, 1 + 0.01WH))||_F,$$

$$P1 = ||H - H^2||_F,\ P2 = ||W - W^2||_F,$$

$$P3 = ||H^\mathsf{T}||_{2,1},\ P4 = ||W^\mathsf{T}W - diag(W^\mathsf{T}W)||_F$$

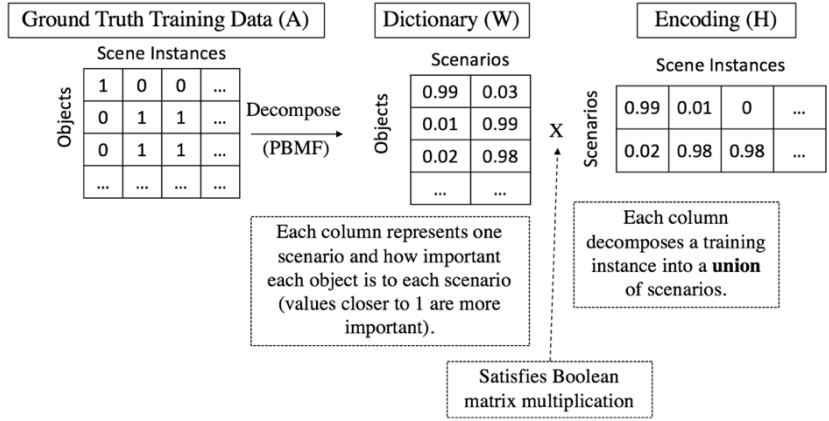

**Fig. 4.** A visual overview of Pseudo-Boolean Matrix Factorization

where $\Omega$ is a weight matrix (based on inverse document weighting from the information retrieval literature) that decreases the importance of common objects and increases the importance of rare objects during the factorization. "$\bullet$" denotes element-wise matrix multiplication. The $\alpha$(s) represent tradeoff parameters which can be set manually, using hyperparameter optimization methods, or automatically using heuristic methods. P0 optimizes the reconstruction error using an approximation of Boolean matrix multiplication. P1 and P2 penalize $W$ and $H$ in order to encourage these matrices to be binary. As with most matrix factorization-based methods for dictionary learning, one needs to select the number of basis vectors (in our case, scenarios). P3 uses the $L_{2,1}$-norm to perform automatic scenario selection to prune the number of scenarios. P4 enforces orthogonality between the scenarios to encourage diversity of the object groupings and minimize redundancy between scenarios. While PBMF is nonconvex, with careful initialization, standard optimization techniques such as gradient descent converge to good local minimums which capture meaningful sets of objects. For example, some of the scenarios learned by our method include: {candelabra, candle, candlestick, fireplace}, {computer, desktop, folder, keyboard, laptop, monitor, mouse, mousepad, printer/copier/scanner, projector, telephone}, and {mirror, sink, soap, soap bar, towel, towel rack}.

In order for scenarios to be useful for our applications, they must satisfy three properties: 1) they must be discriminative for the scene classification task; 2) they must be able to be recognized from visual data; and 3) they must make sense to humans. We experimentally validate each of these claims.

First, we consider how *discriminative scenarios* are for scene classification compared to using object presence. Consider Table 3 that highlights that scenarios, despite being a compressed representation of the object information in a scene, is competitive with using raw object presence features for the scene classification task in terms of classification accuracy. We explore using the "ground truth" scenario data (*i.e.,* the scenarios learned by applying PBMF to the ground truth object data), for 30 scenarios. The 30-dimensional scenario representation is a much more concise representation compared to the 201-dimensional object presence representation. We see that in terms of single-view accuracy, little discriminative information is lost (~3%). Unfortunately, more information is lost in terms of multi-view accuracy (~10-11%). However, Fig. 5 shows that ScenarioNet can recognize scenarios with much higher accuracy than individual objects, and later by comparing predicted scenarios and predicted objects as features for scene classification, the predicted scenario representation is more robust.

**Table 3.** Comparison of scenarios vs. raw object presence features for scene classification accuracy.

| Method | Single-View Accuracy | All-View Accuracy |
|---|---|---|
| Ground Truth Object Presence + Logistic Regression | 0.654 | 0.793 |
| Ground Truth Scenarios + Logistic Regression | 0.620 | 0.684 |

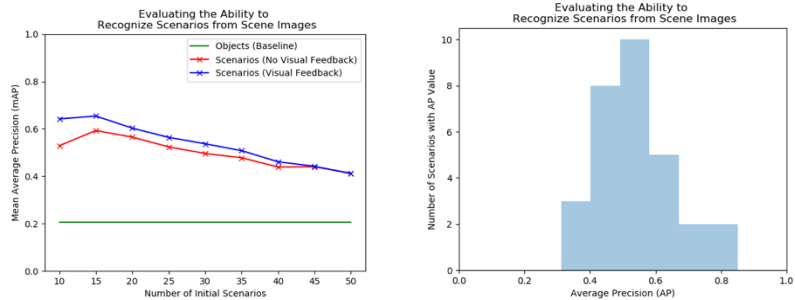

**Fig. 5.** Left: Understanding how scenario recognition performance is affected by the number of scenarios. Also, refining the dictionary based on visual feedback is useful for improving scenario recognition. Right: We train a model to learn to recognize 30 scenarios. Results show that scenarios can be relatively accurately recognized from scene images.

We want to examine if the learned scenarios *make sense to humans* (*i.e.,* do the object groupings make sense?) and if scenario-based explanations for scene classification make sense to humans. In one experiment, we presented 30 scenarios (learned by our model with visual feedback) to 20 English-speaking participants using Amazon's Mechanical Turk (AMT) service. We asked participants to select whether each scenario is one of the following: "is a meaningful group of objects", "might be a meaningful group of objects but doesn't align with my expectations", "is a meaningless group of objects", or "consists of objects I'm not familiar with". After taking the modal response for each question, respondents found *74.1% of the scenarios were meaningful, 11.1% might be meaningful, and 14.8% were meaningless*. Upon examining the meaningless scenarios, we found that these were often considered "meaningless" because sometimes the PBMF would add one or two seemingly random objects to a meaningful scenario or accidentally merge two meaningful scenarios. These problems may possibly be solved by using a slightly larger number of scenarios. We then evaluated if humans could accurately identify the category of a scene when presented with only the most influential scenarios and their influence score (the predicted scenario probability output by the CNN multiplied by the corresponding weight in the logistic regression model). For this experiment, scenarios were pooled over all views. We gathered 15 English-speaking participants using AMT and for 50 random test scenes that were correctly classified by our model, the participants were given a list of all scenarios with an influence score greater than one and asked to predict the scene class from four choices (one true, three randomly chosen). After taking the modal response for each question, *98% of the scene classifications were correct*, suggesting that the model output plausible explanations.

### 5.3.2 Deep Neural Networks for Mapping from Pixels to Scenarios

A key element of ScenarioNet in mapping pixels to scenarios. We explore two ways of predicting the scenario encodings $H$ for each image. In the first method, ScenarioNet learns $W$ and $H$ using P-BMF on the ground truth object data, using a threshold for $H$ at a value of 0.5 (where a scenario $i$ is considered present in image $j$ if its encoding value $H_{ij}$ is greater than the threshold), and then train a standard ResNet-18 CNN model to perform multi-label recognition (using weighted binary multi-label cross-entropy as the loss function). In this case, the dictionary is always held static and does not receive any feedback from the visual data. In the second method, in order to incorporate visual feedback into the dictionary, the dictionary is refined from the ResNet-18 training to predict $H \geq 0.5$:

1. Perform PBMF to obtain an initial dictionary $W^{(0)}$ and ground truth scenarios $H^{(0)}$.
2. Prune the scenarios based on the $L_{2,1}$-norm of $H^{(0)T}$ (the transpose of $H^{(0)}$).
3. Threshold $H^{(t)} \geq 0.5$.
4. Train CNN to estimate scenario presence from images.
5. Extract the predicted scenario probabilities $\widehat{H}^{(t)}$ from all training examples.
6. Refine the dictionary by holding $\widehat{H}^{(t)}$ constant and solving for $W^{(t)}$.
7. Get new ground truth scenarios by holding $W^{(t)}$ constant and solving for $H^{(t+1)}$.
8. Repeat 2-7 until the stopping criteria on the validation data is met.

Fig. 6 shows an example output of our "ScenarioNet" model applied to an image from the ADE20K dataset [14].

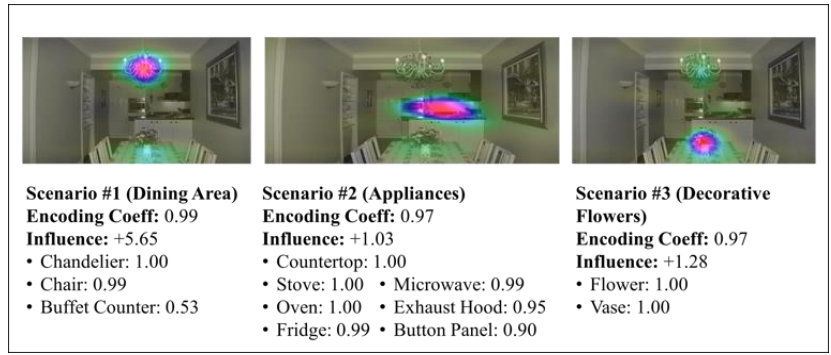

**Fig. 6.** An example output of the ScenarioNet model showing the top-3 predicted scenarios for a dining room scene, and apply the class activation mapping [76] technique to highlight where the net is attending when predicting each scenario. Image taken with permission from [84].

For the baseline approach we use a ResNet-18 CNN [128] that is pre-trained on the Places-365 scene classification dataset [10] and fine-tuned on individual images from our dataset. The CNN is trained for a maximum of 100 epochs (training iterations through the entire dataset) with early stopping based on the validation data. Typically the model converges in fewer than ten epochs. A batch size of 16, learning rate 1.0e-4, and a weight decay of 1.0e-5 are used. The AMSGRAD optimizer [129] is used. The learning rate is reduced when the training loss plateaus.

### 5.3.3 High-Level Information Fusion Using Scenarios

The goal of our approach is to provide a human-understandable representation (scenarios) that enables the ScenarioNet model to output decisions with explanations, utilizing a mapping from pixels to these representations using neural networks. Finally, a concern is how information can be fused as new views are encountered. With scenarios, incorporating new information is simple. If a scenario is detected in a view, it should have a value close to one (*e.g.,* high correlation) and zero otherwise. This enables ScenarioNet to perform high-level information fusing between views using a very straighforward process: ScenarioNet simply needs to take the maximum value for each scenario over all of the views encountered up to this point. It should be noted that there are likely improved fusion schemes that are not as susceptible to noise in the scenario recognition (*i.e.,* because scenarios can be incorrectly predicted from images), but our experiments show that this *simple max-pooling* scheme generally works well.

We can now evaluate how powerful a scenario-based representation is for traditional scene classification. From Table 4, we see that by using noisy object and scenario predictions, ScenarioNet does lose a noticeable amount of predictive performance compared to using ground truth object information; however,  the ScenarioNet representation outperforms the standard ResNet-18 CNN model, and scenarios are much easier to recognize than individual objects, so our explanations are much more trustworthy.

**Table 4.** Understanding the discriminative power of predicted scenarios as a representation for scene classification.

| Method | Single-View Accuracy | All-View Accuracy |
|---|---|---|
| Ground Truth Scenarios + Logistic Regression(30 Scenarios) | 0.620 | 0.684 |
| Standard ResNet-18 CNN | 0.504 | 0.586 |
| Predicted Object Probabilities + Logistic Regression | 0.473 | 0.564 |
| Predicted Scenario Scores + Logistic Regression(30 Scenarios) | 0.476 | 0.607 |

### 5.4 Decision Making and Prediction

Once we have the scenarios for a given view or sequence of views, we need to decide whether: 1) to assign with high confidence some label to the scene (*e.g.,* the category of the scene or actions taking place in the scene), 2) to determine that the scene is atypical given the agent/model's past experiences, or 3) to acknowledge that no prediction can be made with high confidence, and as a result, the sensor must be directed/adjusted to collect additional information/evidence for making a decision (via directives from the model, discussed in Section 5.5.2). In this section, we will focus on constructing a model to address the first two points.

For ScenarioNet to be interpretable when making a prediction, scenarios need to be composed as human-understandable features with an interpretable classifier. The simplest choice for the classifier is a linear multinomial logistic regression model. However, logisitic regression is not designed to be compatible with open set classification. Instead, we will use a *Weibull-Calibrated Support Vector Machine (W-SVM)* [53]. The W-SVM is an extension of the popular support vector machine (SVM) [130] which is capable of performing multi-class classification and open set recognition. The W-SVM is based on extreme value theory, which enables the model to reject assigning a label when an unknown scene is encountered despite never seeing an example of this scene category in the training data. This is in contrast to traditional closed set classifiers which are forced to always output a prediction from the set of known labels, and often will output predictions with high confidence even when faced with out-of-distribution data because they are not properly calibrated (*e.g.,* see [131] for a discussion on the necessity of calibrating deep neural networks).

The W-SVM formulation makes use of two types of SVMs: a one-class SVM (OC-SVM) [132] and a one-vs-rest SVM, both trained for each of the known classes. For each of these models, Weibull distributions are fit based on the distances between a training sample (in feature space) and each decision boundary. The OC-SVM enables computing the probabilitity of inclusion for a given class $P_O(y|f(x))$. The one-vs-rest SVMs fits two Weibull distributions. The first distribution computes the probability for inclusion in the target class:

$$P_{R+}(y|f(x)) = 1 - e^{-\left(\frac{f(x)-v_{R+}}{\gamma_{R+}}\right)^{\kappa_{R+}}} \tag{1}$$

where f(x) is the data, y is the class, and v, γ, and κ are, respectively, the location, scale, and shape parameters of the Weibull distribution. The second distribution is a reverse Weibull distribution which enables us to compute the probability that the data doesn't belong to one of the other known classes:

$$P_{R-}(y|f(x)) = e^{-\left(\frac{f(x)-v_{R-}}{\gamma_{R-}}\right)^{\kappa_{R-}}} \tag{2}$$

Parameters of the various Weibull distributions are fit using maximum likelihood estimation.

In our implementation, an RBF kernel is used for the OC-SVM, and a linear kernel is used for the one-vs-rest SVM, affording an easy-to-interpret classifier in combination with understandable features, resulting in an explainable model. At test time, to determine if a new data point belongs to one of the known classes, we run the following procedure:

1. Test $P_O(y|f(x)) > \delta_O$ for each class y where $\delta_O$ is a small threshold. If no class satisfies this condition, then the model can reject making a class assignment and say the scene is atypical.
2. For those classes that pass the first test, then perform a second test. We test $P_{R+}(y|f(x)) * P_{R-}(y|f(x)) > \delta_R$ where $\delta_R$ is some threshold. This tests the probability that the input is from the positive class and not from any of the known negative classes. If no class satisfies this condition, then we can reject making a class assignment and declare the scene is atypical.
3. For the remaining classes, we select $y^*$ as our prediction, where $y^*$ is the argmax of $P_{R+}(y|f(x)) * P_{R-}(y|f(x))$.

The thresholds $\delta_O$ and $\delta_R$ are determined empirically via cross-validation. We can also compute class-specific probabilities by computing $P_{R+}(y|f(x)) * P_{R-}(y|f(x)) * I(P_O(y|f(x)) > \delta_O)$ for each y where I is the indicator function. These probabilities are useful when incorporating exploration into our method in Section 5.5.2.

We now experimentally validate the performance of the W-SVM classifier for the task of scene classification. Table 5 shows that W-SVM is competitive (in terms of single-view and all-view accuracy) with logistic regression when all classes are known. We also compared W-SVM to other open set classification models, including the nearest neighbor + compact abating probability (NN-CAP) model [53], "probability of inclusion" support vector machine (PI-SVM) [52], extreme value machine (EVM) [55], and openmax classifier [54], on the open set scene classification task. In this setting, we trained on half of the classes and tested on all classes, repeating this experiment for ten random trials. Table 6 highlights that the W-SVM is relatively effective for the open set scene classification task. The W-SVM was competitive with the other open set classification models, and compared to these other models, W-SVM does a good job balancing performance on the known class multi-class classification task and unknown class rejection task.

**Table 5.** Performance of logistic regression and W-SVM (using 30 predicted scenario probabilities as features) when all classes are known.

| Method | Single-View Accuracy | All-View Accuracy |
|---|---|---|
| Predicted Scenario Scores + Logistic Regression | 0.476 | 0.607 |
| Predicted Scenario Scores + W-SVM | 0.459 | 0.593 |

**Table 6.** Performance of combining scenarios with W-SVM on open set recognition using 30 scenarios and all views with 7 known classes and 7 unknown classes. Results are averaged over 10 random trials.

| Method | Known Class Accuracy | Unknown Class Precision | Unknown Class Recall | Unknown Class AUPRC |
|---|---|---|---|---|
| Predicted Scenario Scores + NN-CAP | 0.424 | 0.489 | 0.581 | 0.563 |
| Predicted Scenario Scores + PI-SVM | 0.511 | 0.594 | 0.613 | 0.594 |
| Predicted Scenario Scores + EVM | 0.513 | 0.542 | 0.632 | 0.615 |
| Predicted Scenario Scores + OpenMax | 0.460 | 0.554 | 0.609 | 0.574 |
| Predicted Scenario Scores + W-SVM | 0.525 | 0.577 | 0.617 | 0.603 |

### 5.5 Updating the Models and Adjusting the Sensors

Finally, we need to discuss how to update the model as new data is encountered and how the model adjusts the sensor (camera) to capture more discriminative portions of the scene.

### 5.5.1 Updating the Models

The chapters has thus far shown that ScenarioNet can recognize when a new scene category is encountered, but ScenarioNet is also capable of updating its internal knowledge (scenarios) and ML models (CNNs for scenario recognition) to account for this new information. In this section, we propose a simple extension to PBMF whereby ScenarioNet learrns to augment the scenario dictionary using only instances from a new scene category. This can be done by solving for a small matrix $W^{(c)}$, representing class-specific scenarios, using only ground truth object data from the new class instances $A^{(c)}$:

$$\min_{W^{(c)}, H^{(c)}} P0 + \alpha_1 * P1 + \alpha_2 * P2 + \alpha_3 * P3 + \alpha_4 * P4$$

$$s.t.\ W^{(new)} = [W, W^{(c)}], W^{(new)} \in [0,1]^{m \times (k+k_c)}, W^{(c)} \in [0,1]^{m \times k_c}, H^{(c)} \in [0,1]^{k_c \times n_c}$$

$$\Omega_{ij} = \max\left(A_{ij}^{(c)} * \left(1 + \log\left(\frac{N_{instances}}{N_{objects}}\right)\right), 0.5\right),$$

$$P0 = ||\Omega \bullet (A^{(c)} - \min(W^{(new)}H^{(c)}, 1 + 0.01W^{(new)}H^{(c)}))||_F,$$

$$P1 = ||H^{(c)} - H^{(c)2}||_F,\ P2 = ||W^{(new)} - W^{(new)2}||_F,$$

$$P3 = ||H^{(c)\intercal}||_{2,1},$$

$$P4 = ||W^{(new)\intercal}W^{(new)} - diag(W^{(new)\intercal}W^{(new)})||_F$$

Every time a new class $c$ is added, ScenarioNet solves for the new class-specific scenarios $W^{(c)}$ and appends them to the old dictionary $W^{(new)} = [W, W^{(c)}]$.

To validate ScenarioNet for dynamically updating the scenario dictionary (Dynamic PBMF), experiments are run over ten trials where an initial dictionary is learned using seven classes and twenty scenarios. For each remaining class, ten class-specific scenarios are learned, pruned based on the $L_{2,1}$-norm of $H^{(c)T}$, and appended to the existing dictionary. For each trial, as a baseline, ScenarioNet learns a set of scenarios using all data for an equal number of scenarios as Dynamic PBMF outputs. We compare Dynamic PBMF to regular PBMF in terms of 1) reconstruction error on the entire dataset and 2) discriminability on both the single-view and all-view scene classification tasks. Results appear in Table 7. On average, Dynamic PBMF learns 43.5 scenarios. As expected, regular PBMF results in a lower reconstruction error and higher scene classification accuracies, but Dynamic PBMF is very competitive. The reconstruction error of Dynamic PBMF is only 1.2 times larger than regular PBMF, and scene classification accuracy is within 1% on the all-view classification task, and within 5% on the single-view classification task.

**Table 7.** Quality comparison of a dictionary learned by Dynamic PBMF to one learned with regular PBMF.

| Method | Single-View Accuracy | All-View Accuracy | Reconstruction Error |
|---|---|---|---|
| Ground Truth Scenarios (Static) + Logistic Regression | 0.620 | 0.684 | 205.2 |
| Ground Truth Scenarios (Dynamic) + Logistic Regression | 0.573 | 0.678 | 247.6 |

The scenario dictionary can be updated in an dynamic manner, but we also need to be able to efficiently update the ML model to recognize these new scenarios from visual data; *i.e.,* we need to update the CNN model to account for the new scenarios instead of retraining the CNN model from scratch using all previously collected data. We propose a branching CNN model: when a new class is encountered, a new set of class-specific scenarios is learned using Dynamic PBMF, and a new branch of our existing CNN is trained to perform multi-scenario recognition on just the new scenarios using only data for the new classes (see Fig. 7). Since these branches are learned on very limited data, they train very fast, but this means they might be prone to overfitting. Interestingly, the results in Table 8 show otherwise. Ten trials were conducted where an initial model was trained on data from half of the scene categories and then a branched model was learned on the remaining seven scene categories. On average, 39.3 scenarios were learned per trial. The branching CNN achieves comparable scenario recognition performance compared to a single model trained on all data at once. The branching model achieves superior single-view scene classification performance and equivalent all-view scene classification performance when compared to the traditional model. These results provide promising evidence that since class-specific scenarios are ideally independent between different scene categories, they can be learned with significantly limited data. It should be noted that the ability of the model to generalize quickly might be due in part because the ResNet is pre-trained on Places-365, a large-scale scene dataset. It should also be noted that the shared layers of the model are only finetuned on the initial seven classes and are not updated as new classes are encountered. In future work, we'd like to explore principled methods for updating these shared layers while minimizing catastrophic forgetting (abruptly forgetting previously learned weight values as the model is updated to account for new tasks/data) [133-135].

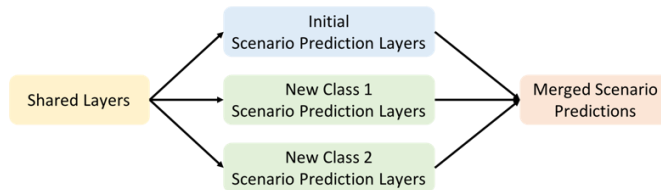

**Fig. 7** A high-level overview of the branching ScenarioNet architecture where as Dynamic PBMF discovers new class-specific scenarios, the scenario-predicting neural network is updated by learning new branches consisting of layers designed to recognize the new class-specific scenarios.

**Table 8.** Understanding the the performance of the branching convolutional neural network model for scenario recognition

| Method | Scenario Recognition mAP | Single-View Accuracy | All-View Accuracy |
|---|---|---|---|
| Predicted Scenario Probabilities (Dynamic PBMF + Traditional Model) + Log. Reg. | 0.414 | 0.503 | 0.649 |
| Predicted Scenario Probabilities (Dynamic PBMF + Branching Model) + Log. Reg. | 0.410 | 0.543 | 0.648 |
| Predicted Scenario Probabilities (Dynamic PBMF + Traditional Model) + W-SVM | 0.414 | 0.488 | 0.587 |
| Predicted Scenario Probabilities (Dynamic PBMF + Branching Model) + W-SVM | 0.410 | 0.525 | 0.634 |

### 5.5.2 Adjusting the Sensors

Finally, adjusting the sensors as new views are encountered would increase robustness. In this section, we model the exploration component of ScenarioNet. We define a Markov Decision Process [136] for the problem of active explanation-driven classification of indoor scenes. We define a *state* to be the vector of class probabilities output by W-SVM concatenated with the rejection score (one minus the maximum probability output by W-SVM), and the number of views seen. There are four *actions*: 1) make a class prediction, 2) reject making any decision and end the exploration process, 3) adjust the camera to the nearest unseen view, and/or 4) adjust the camera to the furthest unseen view. We define the *rewards* as: -1 if the view is changed when the agent would have made the correct prediction, -8 if the model predicts an incorrect class, -8 if the model refuses to make a prediction and ends exploration when it would have predicted the correct class, and 8 + (number of remaining unseen views)$^\psi$ if a correct classification is made. $\psi$ is a parameter which controls the trade-off between accuracy and exploration. The *terminal states* are when either a prediction is made or the agent rejects making a prediction and ends exploration. We use linear function approximation of the Q-value with experience replay as our reinforcement learning algorithm.

We conduct experiments to validate our approach. 10 trials are run. In each trial, 7 known classes and 7 unknown classes are selected randomly. A single W-SVM model is trained for 1-8 views, randomly sampled. We measure the average number of actions taken, the accuracy of the predictions on the known classes, and the precision and recall of rejecting the unknown classes (see Table 9). Compared to the results in Table 6, which involved no active exploration, if we set $\psi = 0$, similar known class accuracy and unknown class precision are achieved while unknown class recall significantly improves, and if we set $\psi = 1.5$, similar known class accuracy, unknown class precision, and unknown class recall are achieved while we only need to consider ~4 views on average, suggesting exploration is useful.

**Table 9.** Results for active explanation-driven classification of indoor scenes

| $\phi$ | Mean Number of Actions | Known Class Accuracy | Unknown Class Recall | Unknown Class Precision |
|---|---|---|---|---|
| 0 | 6.66 | 0.49 | 0.78 | 0.61 |
| 1.5 | 4.19 | 0.46 | 0.56 | 0.59 |

## 6 Remaining Challenges and Future Work

We have discussed our general DDDAS-based ScenarioNet framework and demonstrated how it can be applied to a simple problem domain. However, many open questions remain. For example, consider the following technical questions:

How well does our approach work when it is applied to the more complicated domains discussed in Section 2? Are there alternative principled and automatic ways to determine human-understandable representations for a given-domain? What are the best methods (ML or otherwise) to map from the sensor space to the semantic representation? What is the best method for fusing information between views? If CNNs are used to map from sensor measurements to semantic features, how do we overcome the problem of catastrophic forgetting? What is the best algorithm for open set recognition for some given feature set? What is the best Markov Decision Process for a given problem, and can this be discovered automatically in an online manner? How much data do we need in order to create an initial representation and prediction model before updating the model via exploration becomes feasible? Instead of updating on a per-class level, can we update in a streaming manner where the model is updated as every new data point is encountered.

It should be noted that even our simple case study (active explanation-driven indoor scene classification) is far from solved, and there is still room for large amounts of improvement. We can also consider more foundational questions relating to determining if the DDDAS ScenarioNet framework is the "best" or most general framework for addressing dynamic data driven explainable scene understanding or if alternatives exist; and we also need to consider correspondence of the use of such a framework for high-impact domains such as defense and surveillance and emerging ethical requirements.

## 7 Summary and Conclusion

We presented a DDDAS framework (ACUMEN) for the active explanation-driven understanding and classification of scenes that utilize the ScenarioNet. Unlike existing approaches for scene understanding, the ACUMEN approach doesn't assume 1) perfectly representative single views of scenes as input, 2) closed set assumptions about the types of scenes that can be encountered, and 3) pre-defined model explainability and interpretability priortizations. By overcoming the limitations of existing methods, the ACUMEN framework enables the next generation of scene understanding and is much more useful in practical/real world settings. In particular, we discussed how the ACUMEN framework can be applied to high-impact domains such as remote sensing, defense, surveillance, and robotics. To demonstrate how the ACUMEN framework can be adapted to a specific domain and to validate the effectiveness and utility of the approaches we presented, we introduced 1) a simple case study, the active explanation-driven classification of indoor scenes, and 2) an instantiation of our framework applied to this case study. The ACUMEN framework is still still being enhanced for real-world robustness of which the chapter provides areas of research extensions such as enhanced contextual models, user active learning, and information fusion methods.

## Acknowledgements

Parts of this work were supported by the DDDAS program of the Air Force Office of Scientific Research (AFOSR) and by the National Science Foundation Graduate Research Fellowship under Grant No. DGE-1433187.